# The Potential of Large Language Models for Improving Probability Learning: A Study on ChatGPT3.5 and First-Year Computer Engineering Students

## Authors

Ángel Udías[1], Antonio Alonso-Ayuso[3], Ignacio Sánchez[2], Sonia Hernández[1], María Eugenia Castellanos[1], Raquel Montes Diez[1], Emilio L. Cano[3]

**Affiliations**

1. Rey Juan Carlos University, Móstoles, Spain
2. European Commission, Joint Research Centre, via E. Fermi 2749, 21020 Ispra (VA), Italy
3. CETINIA: Research Centre for Intelligent Information Technologies (CETINIA-DSLAB), Rey Juan Carlos University, 28933 Móstoles, Spain

corresponding author: Angel Udias (angelluis.udias@urjc.es)

Keywords: ChatGPT3.5, Large language Models (LLMs), Probability Problems, Artificial Intelligence in Education

## Abstract

In this paper, we assess the efficacy of ChatGPT (version Feb 2023), a large-scale language model, in solving probability problems typically presented in introductory computer engineering exams. Our study comprised a set of 23 probability exercises administered to students at Rey Juan Carlos University (URJC) in Madrid. The responses produced by ChatGPT were evaluated by a group of five statistics professors, who assessed them qualitatively and assigned grades based on the same criteria used for students.

Our results indicate that ChatGPT surpasses the average student in terms of phrasing, organization, and logical reasoning. The model's performance remained consistent for both the Spanish and English versions of the exercises. However, ChatGPT encountered difficulties in executing basic numerical operations. Our experiments demonstrate that requesting ChatGPT to provide the solution in the form of an R script proved to be an effective approach for overcoming these limitations.

In summary, our results indicate that ChatGPT surpasses the average student in solving probability problems commonly presented in introductory computer engineering exams. Nonetheless, the model exhibits limitations in reasoning around certain probability concepts. The model's ability to deliver high-quality explanations and illustrate solutions in any programming language, coupled with its performance in solving probability exercises, suggests that large language models have the potential to serve as learning assistants.

## BACKGROUND & SUMMARY

Teaching has undergone significant changes in recent years, with a particular emphasis on digitalization and the use of computer tools (Hofer et al., 2021). This shift, prompted by the pandemic, has accelerated the adoption of technology in education, with the development of audio-visual materials and automatically grading exercises, in addition to videoconferencing. These tools have enabled educators to provide more interactive and engaging learning experiences, allowing students to explore concepts more deeply and at their own pace (Giesbers et al., 2013).

However, it is important to note that while these developments have certainly improved the learning experience, the importance of knowledge consolidation exercises, particularly in mathematics and statistics. cannot be overstated. These exercises are crucial in ensuring that students not only understand the concepts but also know how to use them to solve real-world problems (Freeman et al., 2014). In fact, research has shown that students who engage in active learning and practice display better retention of information and perform more effectively on exams on exams (Carini et al., 2006).

Furthermore, the utilization of technology in education has presented new challenges. For instance, students may encounter difficulties staying engaged and motivated in online or hybrid learning environments, resulting in higher dropout rates. Additionally, there may be concerns regarding the quality of online resources and the dependability of automated grading systems. Despite these obstacles, the integration of technology in education is here to stay, and educators must continue to adapt to these changes to ensure that students are adequately prepared for the future. This includes integrating innovative tools such as artificial intelligence (Guan et al., 2020) or chatbots, which can provide students with immediate feedback and support (Wollny et al., 2021).

Translating real-world problems presented in natural language into abstract problems expressed in variables, parameters, functions, and equations is one of the most significant challenges faced by statistics students (Garfield et al., 2008). For example, many students may find it difficult to identify a binomial distribution when inspecting a batch of parts and discarding it if the number of defective parts exceeds a specific threshold. Additionally, statistics is a cross-disciplinary field relevant to various disciplines, including economics, computer science, natural sciences, medicine, and linguistics.

Therefore, it is crucial to provide students with a diverse array of exercises that challenge them to identify the most suitable statistical or probability model to represent a given situation and to apply appropriate theoretical tools accordingly.

Large language models (LLMs) have the potential to bring about a revolution in the field of education (Floridi & Chiriatti, 2020). These models rely on transformer architectures (Vaswani et al., 2017) that enable them to process natural language input and follow instructions through Reinforcement Learning based on Human Feedback (RLHF) (Ouyang et al., 2022). LLMs can also be fine-tuned to specific contexts and aligned to generate human-like responses with impressive accuracy (Brown et al., 2020) while avoiding the production of irrelevant or harmful content.

Generative pre-Trained Transformers (GPT) are a family of LLMs introduced by OpenAI in 2018 (Radford et al., 2018). Over the various generations of GPT models, their size (i.e., number of parameters) and complexity have substantially increased. The third-generation models have 175 billion parameters and are fine-tuned to follow instructions via RLHF (Ouyang et al., 2022). ChatGPT, which was released in November 2022, is based on the GPT 3.5 model, with additional fine-tuning through supervised learning and RLHF to enable interaction with users via a text dialog. ChatGPT is available to users through a web interface, and the underlying model is regularly updated. The latest version available as of writing this paper is February 13, 2023

LLMs have a wide range of potential applications across various domains. In the field of education, their possible uses are diverse and extensive, including but not limited to automated essay grading, personalized learning, intelligent tutoring systems, and conversational agents (Holmes & Tuomi, 2022).

Despite the potential benefits of using large language models in education, their integration also presents significant challenges due to their current limitations. One of the most pressing issues is the potential for inaccuracies, which can lead to the models presenting compelling yet false information to the user. Additionally, biases in the output of these models can perpetuate and even amplify existing societal inequalities, posing a significant concern (Shahriar & Hayawi, 2022; Barikeri et al., 2021).

To effectively address these challenges, it is essential for teachers and learners to develop a set of competencies and literacies that enable them to understand the technology and its limitations. This includes critical thinking skills and strategies for fact-checking, as well as an understanding of the potential biases and risks associated with large language models (Bender & Friedman, 2020). In conclusion, while the potential benefits of large language models in education are significant, it is crucial to approach their integration with a clear pedagogical strategy and a strong focus on responsible and ethical use (Kasneci et al., 2023; Holmes et al., 2022).

The objective of this research is two-fold. First, we evaluate the capability of ChatGPT (version Feb 2023) to solve probability exercises commonly administered in first-year computer engineering exams, and to compare the answers generated by the model with those evaluated by expert professors. Second, we provide a qualitative assessment of the results obtained and discuss the potential application of Large Language Models in education at university level.

The paper is structured as follows: Section 2 provides a detailed description of the experiment, including the methodology used. Section 3 is dedicated to the analysis of the results, and Section 4 presents the conclusions and suggestions for future research.

## METHODS

To conduct the experiment, the first step is to gather a representative set of probability exercises. We have compiled 23 exercises that were initially proposed to first-year Computer Engineering students enrolled in introductory courses of statistics at Rey Juan Carlos University in Madrid, Spain. The URJC degree's cut-off mark, like that of other Spanish universities, is determined by the last student's mark admitted to each degree annually, which varies based on applicants' access marks and the number of places offered. Specifically, for the Computer Engineering degree at URJC, the cut-off marks were 7.371, 7.529, and 7.543 out of 10 points for the years 2020, 2021, and 2022, respectively.

The exercises were originally created by six lecturers specifically for their respective examinations and have not been shared on the internet or made publicly available. Therefore, it is unlikely, though not impossible, that they have been included in an AI training database. It is possible that similar but not identical exercises may exist. The exercises have been systematically categorized into one or two of the nine distinct categories based on the unique attributes of the questions posed (see Table 1). Additionally, each exercise may consist of one or more questions.

The wording (in Spanish) for each exercise can be found in the additional materials. We also gathered data on the students' grades for each group in which the exercises were administered, as shown in Table 1. For each exercise, the following information is provided: category, number of sections, number of students with available results, average score (out of 10), standard deviation, and number of students who received the minimum score (0) and the maximum score (10).

*Table 1: Summary information for each exercise. Categorisation of exercises: CB: Combinatorial; CD: Conditional; BY: Bayes; IT: Intersection; DF: Distribution functions; BN: Binomial; NO: Normal; PS: Poisson; GE: Geometric.*

| ID | Category | Number of questions | Number of students | Mean | Std. dev. | % of Zeros | % of Tens |
|---|---|---|---|---|---|---|---|
| 1 | CB | 1 | 65 | 4.41 | 4.87 | 53.8 | 40.0 |
| 2 | CB, BY | 1 | 32 | 4.42 | 4.45 | 43.8 | 31.3 |

| 3  | CB     | 2 | 47  | 6.52 | 3.40 | 12.8 | 34.0 |
|----|--------|---|-----|------|------|------|------|
| 4  | CB, CD | 3 | 68  | 5.25 | 4.16 | 29.4 | 27.9 |
| 5  | CD, BY | 3 | 54  | 6.17 | 3.69 | 7.4  | 38.9 |
| 6  | CD, IT | 5 | 126 | 5.39 | 3.47 | 9.5  | 22.2 |
| 7  | CB     | 2 | 68  | 5.25 | 4.16 | 29.4 | 27.9 |
| 8  | CD, BY | 3 | 76  | 7.44 | 3.43 | 9.2  | 55.3 |
| 9  | BY     | 3 | 55  | 7.42 | 3.41 | 9.1  | 54.5 |
| 10 | CB, CD | 3 | 55  | 2.97 | 2.85 | 21.8 | 5.5  |
| 11 | CD, BY | 2 | 49  | 7.00 | 3.20 | 10.2 | 28.6 |
| 12 | BY     | 2 | 49  | 6.15 | 3.84 | 18.4 | 38.8 |
| 13 | CD     | 3 | 80  | 5.08 | 3.82 | 23.8 | 26.3 |
| 14 | BY     | 3 | 14  | 5.95 | 2.16 | 0.0  | 14.3 |
| 15 | DF, CD | 4 | 30  | 4.94 | 3.18 | 16.7 | 3.3  |
| 16 | DF, BN | 4 | 47  | 4.81 | 3.76 | 25.5 | 25.5 |
| 17 | DF, BN | 5 | 50  | 6.71 | 3.23 | 12.0 | 16.0 |
| 18 | NO     | 2 | 67  | 4.34 | 3.50 | 17.9 | 7.5  |
| 19 | NO     | 1 | 77  | 4.92 | 4.45 | 36.4 | 36.4 |
| 20 | NO, CD | 3 | 49  | 6.29 | 2.95 | 4.1  | 28.6 |
| 21 | BN, GE | 5 | 49  | 8.10 | 2.15 | 0.0  | 44.9 |
| 22 | PS, GE | 4 | 76  | 5.59 | 3.53 | 0.0  | 26.3 |
| 23 | PS     | 4 | 76  | 3.93 | 2.75 | 15.8 | 5.3  |

In the second step, the 23 exercises were presented to ChatGPT3.5 (Feb 13 version) via the web interface (https://chat.openai.com/chat), and the responses were recorded. The exercises were provided in their original form, as written by the lecturers in the Spanish language, without any modifications or clarifications. In the case where an exercise had multiple questions, each question was presented in a new chat along with the main statement. The 23 exercises included a total of 69 questions.

It is worth noting that ChatGPT does not always generate the same response to a given exercise. As a result, all exercises were presented to ChatGPT3.5 at least three times on different days between February 13 and March 5, 2023, always by the same person using the same account. The prompts utilized with ChatGPT (Kojima et al., 2032) were created by either directly using the original exercise description or appending it with one of the following sentences:

- "Give me a solution with a brief justification".
- "Give me a solution being concise in the answer".
- "Solve the following exercise by being concise in your explanations".

These three slightly different prompt variations allowed us to explore how ChatGPT can be guided to make the style of its answers similar to those of a university student answering an exam.

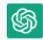

*Figure 1: Screenshot with an image showing the exercise number 1 posed to the GPT and the answer it returns.*

After collecting the different chat responses to each exercise, the analysis process began. It was noticed that ChatGPT generated distinct responses for the three prompt variations employed, although with consistent reasoning. Specifically, for 16 of the exercises, all three responses were deemed equivalent in terms of their reasoning. However, for the remaining 7 exercises, ChatGPT used similar reasoning in two of the responses, and different reasoning in the third. In such cases, both alternative answers were chosen for further evaluation, irrespective of whether the reasoning was correct or the numerical outcome of the exercise. The selection was solely based on the difference between the reasoning applied. This process resulted in a total of 30 answers for the 23 exercises: sixteen exercises had one answer, and the remaining seven had two. It is worth noting that the task of inputting the questions into the chat and selecting the answers was performed by the same lecturer who was not involved in the grading phase.

The next phase of the work involved the evaluation of the ChatGPT responses to each of the 23 exercises. Five lecturers, all of whom had extensive experience in teaching the subject, participated in this phase, and each exercise was evaluated by at least three of them. The lecturers were asked to assess each exercise on a scale of 0 to 10, using the same evaluation criteria they applied to their own students, even if the exercise they were evaluating was not one they had proposed. Along with assigning a final score for the exercise, they were also asked to provide additional comments. These comments included the reason for the main penalty they assigned to the answer (e.g. identification of the problem, explanation of the reasoning, calculations), the level of difficulty of each exercise (low, medium, high), their perception of whether the answer was given by a human or not, and, in the case of exercises with multiple questions, which one is the most penalised.

It is becoming increasingly common to evaluate students' knowledge through self-correction exercises found on platforms such as Moodle (Fatmi et al. 2021; Simanullang et al. 2020). These exercises are typically based on numerical results or the selection of alternatives (Blanco & Ginovart, 2012; Handayanto et al. 2018), and the reasoning used to obtain the result is not assessed, only the final answer. To address this limitation, an additional experiment was conducted by presenting each exercise to ChatGPT again, with the addition of the prompt "Write a code in R language to calculate the probability" before the section of each exercise that asks for a "probability calculation". This allowed ChatGPT to generate R code to express the solution and overcome its documented limitations in performing numerical operations. The R code generated by ChatGPT was executed, and the obtained result was compared with the correct value.

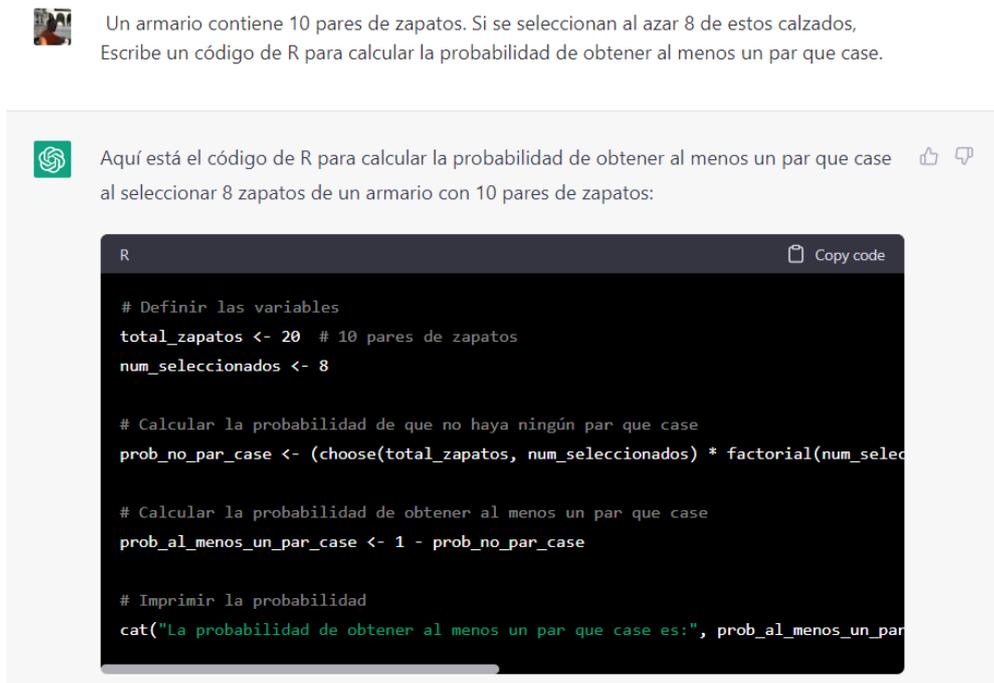

*Figure 2: Illustration capturing the first exercise posed to GPT, where R code is requested and the corresponding response it generates.*

## RESULTS

The number of students who participated in each exercise varied depending on the group or groups assigned the exercise, with the lowest and highest number of students answering an exercise being 14 and 126, respectively (as shown in Table 1). Ideally, the results of each probability problem should reflect the difficulty of the exercise and the level of the classes to which the exercise was given. However, since the entrance grades of students were similar across all groups and the statistics subject encompassing these exam exercises is part of the first-year curriculum, the levels of the students can be considered comparable across all groups.

As mentioned previously, all exercises were assessed by several lecturers (at least three), who were aware that they were evaluating non-human responses. The lecturers' evaluations showed a maximum difference of less than +/-1 point in all ratings, except for exercise number 12, where the two furthest scores differed by +/-2 points. It should be noted that the lecturers exhibited a high level of agreement on the sections where errors were made and the types of errors made. The differences in scores were primarily due to the varying penalty criteria applied by each lecturer in relation to numerical calculation errors. Table 2 summarizes the average scores obtained by the students and the average evaluation of the ChatGPT responses by the lecturers. For the seven exercises in which ChatGPT applied different resolution methods, each answer was evaluated independently, and the results are presented in two rows in Table 2.

In addition, among the seven exercises in which ChatGPT applied two different resolution methodologies at different times, two of the alternative answers (exercises 1 and 19) were entirely incorrect and scored zero in all lecturers' evaluations. It is worth noting that both exercises only have one question. Among the remaining five exercises with two alternative answers, score discrepancies between the two alternatives ranged from 3 points (exercise 11) to a mere 1.2 points (exercise 6). As these exercises have multiple sections, smaller score differences typically indicate errors in some but not all sections.

*Table 2: Mean score of GPT responses (1) and mean score of student responses (2) for each exercise. When there are two answers to the same question, they are displayed in consecutive lines by identifying them with -1 or -2 after the exercise number. Columns from a) to e) correspond to each of the questions in an exercise. Type of errors that lecturers detected in the GPT responses (MA: Meaningless answer; II: Incorrect identification of the problem type; WR: Wrong reasoning; NE: numerical error)*

| Id | (1) | (2) | \multicolumn{5}{c}{Questions penalization} |
|---|---|---|---|---|---|---|---|
| | | | a | b | c | d | e |
| 1-1 | 0.00 | 4.41 | WR | | | | |
| 1-2 | 7.67 | | NE | | | | |
| 2-1 | 5.17 | 4.42 | NE | | | | |
| 2-2 | 7.67 | | NE | | | | |
| 3 | 4.58 | 6.52 | | II | | | |
| 4-1 | 8.33 | 5.25 | | NE | NE | | |
| 4-2 | 9.17 | | | | NE | | |
| 5 | 7.67 | 6.17 | NE | NE | NE | | |
| 6-1 | 7.33 | 5.40 | WR | | | | |
| 6-2 | 8.54 | | | | | NE | NE |
| **7** | 7.59 | 5.25 | | NE | | | |
| **8** | 10.00 | 7.44 | | | | | |
| **9** | 7.94 | 7.42 | | NE | NE | | |
| **10** | 2.75 | 2.97 | | WR | WR | | |
| **11-1** | 3.27 | 7.00 | WR | NE | | | |
| **11-2** | 6.25 | | NE | NE | | | |
| **12** | 6.44 | 6.15 | NE | NE | | | |
| **13** | 5.31 | 5.08 | NE | NE | NE | | |
| **14-1** | 6.11 | 5.95 | NE | NE | NE | | |
| **14-2** | 7.77 | | NE | NE | NE | | |
| **15** | 9.39 | 4.94 | | | NE | NE | |
| **16** | 4.04 | 4.81 | | WR | WR | WR | |
| **17** | 5.25 | 6.71 | MA | WR | WR | | |
| **18** | 8.17 | 4.34 | NE | | | | |
| **19-1** | 0.00 | 4.92 | WR | | | | |
| **19-2** | 7.50 | | NE | | | | |
| **20** | 6.67 | 6.28 | | | WR | | |
| **21** | 9.50 | 8.09 | | | | | NE |
| **22** | 6.61 | 5.59 | | | WR | | |
| **23** | 6.16 | 3.93 | | | NE | | |

Figure 3 displays the distribution of students' marks for each exercise in the form of a violin plot (with a box plot inside). Violin plots offer a more detailed representation than traditional box plots of the distribution of students' marks when the distributions are multimodal and have many observations at the extremes (Hintze & Nelson, 1998). In this case, the colour coding indicates the level of difficulty of each exercise, as assessed by the lecturers. Most of the lecturers (although not all) identified exercises 5, 6, 7, 8, 9, 11, 12, 18, and 20 (green in Figure 3) as easy, and exercises 1, 2, 17, and 19 (pink in Figure 3) as difficult.

Additionally, Figure 3 reveals that the exercises with the highest scores, in decreasing order, were exercises 9, 8, 21, 11, 17, 3, and 20, whereas exercises 18, 23, 10, and 1 received the lowest scores. These results demonstrate a reasonable level of agreement with the lecturers' assessments, except for exercise 17, which yielded satisfactory marks despite being classified as difficult, and exercise 18, where students received low scores despite it being labelled as easy according to the lecturers' assessment.

In exercises with fewer sections or questions, scores tend to be either very high or very low, leading to a more bimodal distribution. For instance, exercises 1, 2, and 19 only have one question, and their distributions are evidently bimodal. The lecturers' average score for evaluating the ChatGPT responses is shown in Figure 3 as a red diamond. In cases where ChatGPT provided two responses for the same exercise, the response with the lowest average score was represented with a red diamond, and the one with the highest score with a blue diamond.

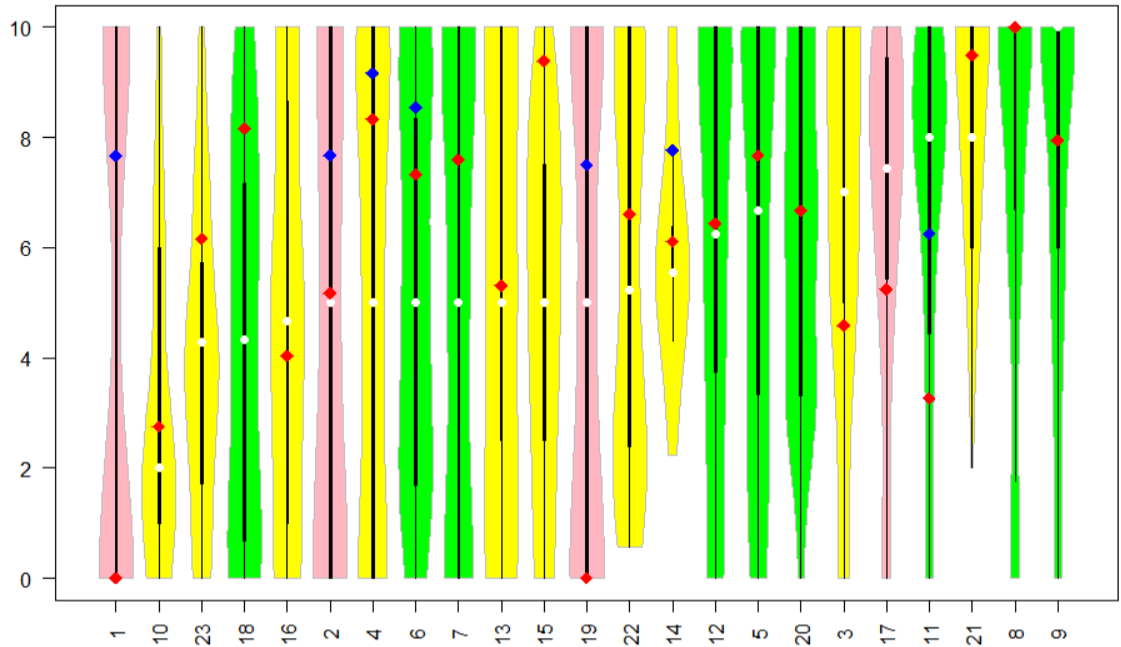

Figure 3: Violin plots with the distribution of student marks sorted by median value. The white dot is the median of the distribution, and the black rectangle represents the interquartile range. The mean of the GPT answer evaluations is shown as a red dot, and a blue dot represents the highest evaluation when there are multiple answers.

Figure 4 compares the mean scores obtained by the students with those generated by ChatGPT for each exercise. In cases where ChatGPT provided two responses, the lower score was selected for analysis. The plot shows data points above and below the diagonal line, indicating exercises where ChatGPT scores were above or below the average of the students, respectively. The results reveal that ChatGPT outperformed the students in 16 out of 23 exercises, which accounts for approximately 70% of the exercises analysed. When only considering the highest score among the two potential ChatGPT responses (Table 2), this percentage increased to 78%.

Out of the 23 exercises analysed, there were six exercises (9, 10, 12, 13, 14, 20) where there were minimal differences between the evaluations of ChatGPT and the students' marks. These exercises required the application of total probability problems or Bayes' theorem, and while ChatGPT's reasoning was correct, its computations were flawed. On the other hand, in six exercises, ChatGPT's average ratings were lower than the students' mean scores, especially in exercises 1 and 19 where ChatGPT received a score of zero from the lecturers. However, in these two exercises, ChatGPT provided alternate responses (as shown in Figure 3) that scored higher than the students' median grades. In the remaining eleven exercises, ChatGPT's response scores were significantly higher than those of the students (as demonstrated in Figure 4). Once again, taking into account the best score in cases with two potential answers, the differences between the students' mean scores and ChatGPT's ratings are even more significant.

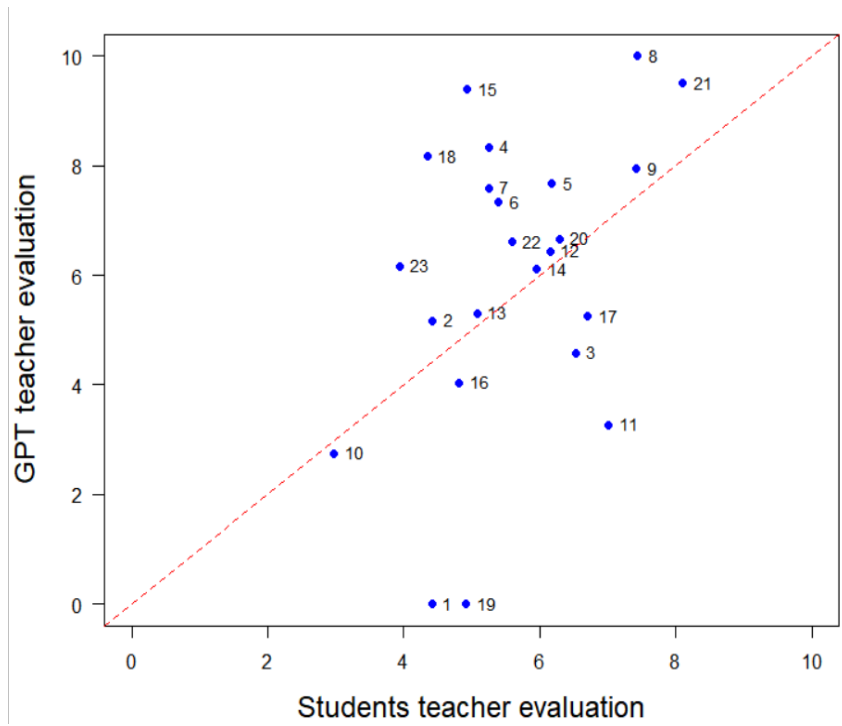

*Figure 4*: Scatter plot comparing the means of students' and GPT scores for each exercise. The red line indicates the 1:1 relationship. The numbers indicate the Id of each exercise.

Despite knowing that they were assessing an AI, lecturers were asked to indicate whether they felt the answers appeared human-like. They could select from the following options: a) Yes, the entire answer; b) Yes, the answer justifies/explains too well; c) Yes, for several reasons; or d) No, not at all. For the 80 % of the exercises, the lecturers felt that the answers did not appear human-like because they "justified/explained too well." Despite the fact that in the multiple-choice GPT exercises each question was posed independently, the lecturers observed that the answers provided using GPT were exceptionally lengthy, surpassing even the extent of detailed responses typically generated by the students themselves. Lecturers were also asked to indicate whether the quality of the explanations was higher or lower than usual for the students' answers, to which the overwhelming majority replied "yes".

The lecturers were also asked to identify the specific types of errors present in each question of every exercise. They were provided with four options: i) Meaningless answer, ii) Incorrect identification of the problem type, iii) Faulty reasoning, or iv) Calculation errors. Table 2 shows the specific errors identified by the lecturers and the corresponding questions where they were found. The evaluation of the 85 questions revealed that 15 of them showed faulty reasoning or incorrect problem identification, while 34 questions contained calculation errors (as shown in Table 2). However, it should be noted that numerical errors were not evaluated in cases where the GPT-generated answers contained faulty reasoning. After the evaluation conducted by the lecturers, we performed a meticulous analysis of all answers, both correct and incorrect, in search of numerical errors. Our findings revealed that out of the 23 exercise answers and 7 alternate exercise answers, only the response to exercise number 8 exhibited a complete absence of numerical errors across all sections (as shown in Table 2). This did not significantly impact the results because the lecturers' evaluation mainly penalized conceptual errors and to a lesser extent, calculation errors. It is important to note that it is unusual for a student to make numerous calculation errors during an exam, whereas the GPT model tends to produce inaccurate answers in this area.

In order to conduct a separate assessment of ChatGPT's reasoning abilities, independent of the limitations it faces in performing numerical operations (Borji, 2023; Frieder et al., 2023), the questions were modified in order to avoid the GPT having to perform numerical calculations. Instead, the questions were rephrased (always in Spanish) to ask the GPT to generate R code (R Core Team, 2022) with which to answer the question. The entire set of questions was presented to the GPT 10 times in a non-consecutive manner for the same person. Subsequently, all the R code generated by the GPT was executed and the numerical output was compared to the correct numerical value for each section question.

*Table 4: Frequency of accurate numerical responses obtained with R- code answers generated by GPT after ten attempts for each section of the 23 exam exercises. Right displaying the final score, calculated as the average of the ten answers provided for each question.*

|     | questions |    |    |    |    |       |
| --- | --- | --- | --- | --- | --- | --- |
| QId | a | b | c | d | e | score |
| 1   | 10 |   |   |   |   | 10.0 |
| 2   | 3  |   |   |   |   | 3.0 |
| 3   | 10 | 8 |   |   |   | 9.0 |
| 4   | 10 | 10 | 0 |   |   | 6.7 |
| 5   | 10 | 10 | 10 |   |   | 10.0 |
| 6   | 3  | 10 | 0 | 1 | 6 | 4.4 |
| 7   | 10 | 0  |   |   |   | 5.0 |
| 8   | 10 | 10 | 0 |   |   | 6.7 |
| 9   | 10 | 10 | 0 |   |   | 6.0 |
| 10  | 10 | 10 | 0 |   |   | 6.0 |
| 11  | 7  | 6  |   |   |   | 6.4 |
| 12  | 5  | 5  |   |   |   | 5.0 |
| 13  | 10 | 10 | 10 |   |   | 10.0 |
| 14  | 10 | 6  | 10 |   |   | 8.2 |
| 15  | 10 | 6  | 1 | 5 |   | 5.5 |
| 16  | 5  | 0  | 0 | 2 | 2 | 1.5 |
| 17  | 3  | 3  | 0 | 0 | 0 | 1.1 |
| 18  | 10 | 10 |   |   |   | 10.0 |
| 19  | 1  |   |   |   |   | 1.0 |
| 20  | 10 | 9  | 0 |   |   | 6.3 |
| 21  | 6  | 10 | 10 | 10 | 10 | 9.2 |
| 22  | 10 | 10 | 5 | 6 |   | 6.1 |
| 23  | 4  | 10 | 6 | 6 |   | 6.0 |

Table 4 presents a summary of the findings from this experiment. Among the 69 questions, ChatGPT was able to consistently provide the accurate numerical answer for 31 questions, whereas for 12 questions, it never produced the precise numerical answer. For the remaining 26 questions, the GPT generated the correct answer with varying frequency, as illustrated in Table 4. The final qualification for each exercise were calculated, considering the value of each question in the exercise, by weighting the scores of all the ten answers provided for each question and the relative value of the question in the exercise (see additional material).

Figure 5 shows a comparison of the distributions of the marks for the 23 exercises of the students' answers and the marks obtained by ChatGPT for both natural language and R-code answers. The results indicate that ChatGPT's responses have a superior mark distribution compared to that of the students, but with greater variability. Specifically, in the natural language response domain, ChatGPT's median mark exceeds the third quartile of the students. It is important to note that the majority of the exercises were not designed specifically for automated grading and involve a scenario where the first section's numerical result carries over to the subsequent sections, which creates an inequitable situation for automated assessment of the exercises. This aspect would undoubtedly penalize not only students but also ChatGPT's R-code assessment.

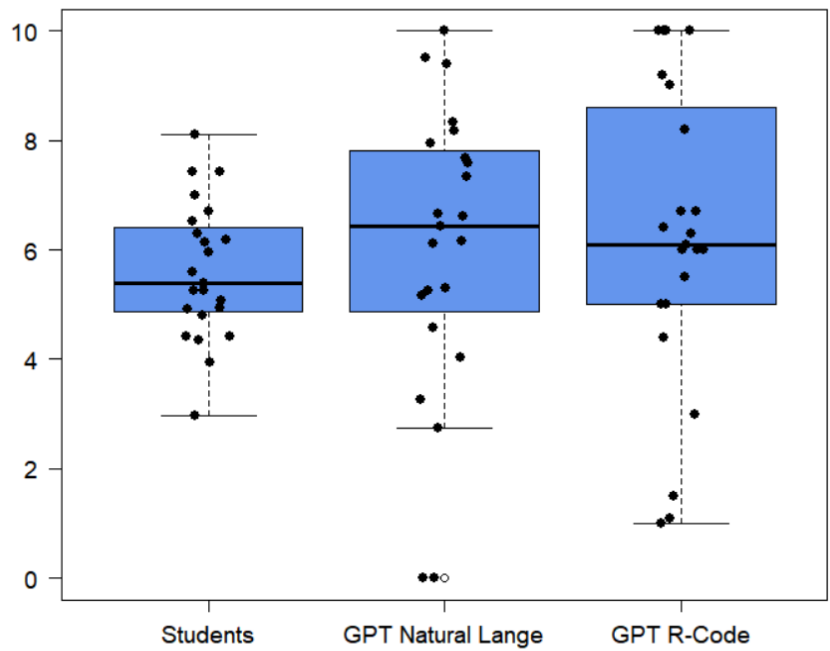

*Figure 5: Box plot with the distribution of marks for all 23 exercises obtained by the students, and the ChatGPT responses in natural language and in R code.*

Figure 6 presents a comparative analysis between the scores obtained by ChatGPT for natural language responses, as evaluated by the lecturers, and R-code responses, as presented in Table 4. In cases where two natural language response alternatives were available, the lower-performing alternative was utilized for the figure. Upon examining the graph, there is not enough evidence to suggest that one problem-solving approach is significantly superior to the other.

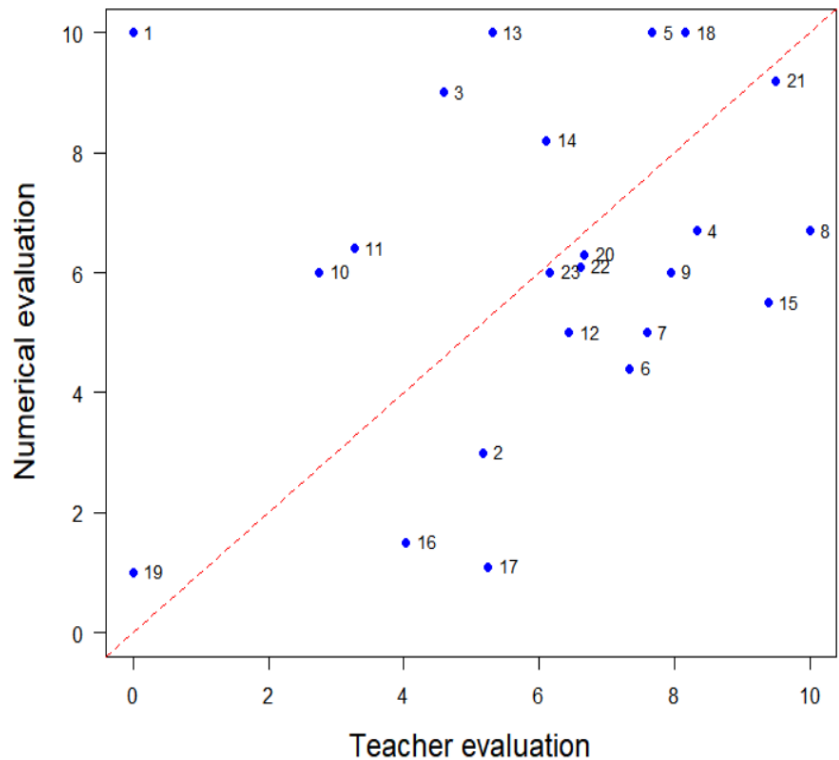

*Figure 6: Scatter plot comparing the mean scores obtained for the ChatGPT in natural language and in R code responses. In the natural language exercises in which two alternative responses are evaluated, only the one with the worst score has been taken into account.*

In the R-coded responses, the problems with numerical calculations disappeared, and the errors were mainly due to conceptual misunderstandings. Notably, for exercises 1, 5, 13, and 18, GPT consistently produced precise results in all ten attempts, indicating the consistent use of the correct methodology. In natural language responses to exercise 1, GPT used the correct methodology in two out of three attempts, with an incorrect methodology applied in one attempt (resulting in a null value in figure 6). In exercises 5, 13 and 18, lecturers penalised answers only for numerical errors (table 2) which are now no longer present.

More striking are the significant improvements observed in the scores obtained for R-code responses to exercises 3, 10, and 11, where lecturers penalized different types of issues in natural language responses other than numerical errors (as shown in table 2). Specifically, in exercise 10, questions b and c, lecturers identified incorrect reasoning in natural language responses (as shown in Table 2), but this issue was resolved in R-code format for question b, while it persisted for question c (Table 4). For question a of exercise 11, reasoning problems only occurred in three out of ten attempts when the response was generated in R-code format. Additionally, in question 3b, the incorrect identification of the problem that appeared in natural language responses was observed in only two out of ten attempts when answered in R-code format (Table 4).

The situation was reversed in exercises 16 and 17, where the evaluations of natural language answers were notably higher than those of R-code responses. In both exercises, the numerical results of all questions depended on the numerical result of the first question. Therefore, R-code responses were heavily penalized if there was any error in the first question.

Of particular interest is the observation that, for certain types of problems, ChatGPT proposes solution methods through R-coded responses that were not presented when the answers were in natural language. Notably, ChatGPT employs techniques such as probability calculations by applying simulation of thousands of repetitions, utilization of R functions to calculate integrals, and employment of R packages or specific or self-written functions to estimate probability, normal or geometric distribution. (Additional material)

Considering both natural language and R-coded answers, we could conclude that the ChatGPT always has conceptual or interpretative errors in less than 15 of the 85 questions. We have considered the possibility that the difficulties in solving these questions properly could be due to a poorer ability to interpret the Spanish language. To verify this, we retested these questions 10 more times in their original Spanish version and their corresponding English translations (additional material) and did not find any difference in the ability to get it right in either language. This seems to indicate that ChatGPT has an equal ability to understand both languages, and that the poor performance in some items is not due to difficulty in interpreting natural language in Spanish.

The additional material describes some of these incorrect answers, and it is worth noting that ChatGPT seems to face difficulties with conditional probability exercises that are not presented in a straightforward manner. While it is able to identify the conditional problem and describe the necessary operations (in both natural language and R-coded answers), it often struggles when performing them. This issue is especially noticeable in the third section of exercises 10, 16, and 20.

## DISCUSSION AND CONCLUSIONS

This study compares the responses to probability exercises generated by ChatGPT with those provided by university students. To achieve this, a representative set of 23 exercises was analyzed. These exercises were previously proposed by five lecturers to computer engineering students taking their first year of statistics courses in a public university in Spain.

Each question was presented to ChatGPT3.5 several times, using exactly the same phrasing in Spanish. Additionally, ChatGPT was asked to generate R code to calculate the numerical value. The written answers were evaluated by the lecturers and compared with those obtained by the students and with the scores of the purely numerical answers of ChatGPT.

ChatGPT demonstrated superior skills in formulation, organization, and reasoning compared to the average student in natural language responses but had frequent errors in numerical operations. When correct reasoning was applied but numerical errors were made, lecturers applied a penalty of approximately 25%. Even so, and considering only the worst answers to each question, ChatGPT outperformed the students in 16 of the 23 exercises.

The results indicate that when ChatGPT provides R-coded responses, the answer is always correct in 45% of the questions, always incorrect in 17%, and correct or incorrect with varying ratios in the remaining 38%. Issues with numerical operations disappear when ChatGPT is asked to generate R-coded answers, and sometimes it employs solution methods (such as simulation of all possibilities and numerical integration) that it did not consider in the answers written in natural language.

The ChatGPT consistently provides unique answers and often uses complex reasoning that is well-articulated and draws on prior knowledge. While the ChatGPT is capable of generating well-founded answers and detailed explanations, there is still room for improvement in its understanding of certain probability concepts. However, when providing answers in R code, the explanations are particularly noteworthy and are a valuable resource for computer engineering students to enhance their coding skills while learning about probability.

Furthermore, the analysis indicates that the ChatGPT's performance in answering questions in both Spanish and English is similar, regardless of whether the answers are provided in natural language or R code.

It is highly likely that future generations of LLMs will effectively overcome the limitations demonstrated by ChatGPT in mathematical operations. Augmented Large Language Models, such as those proposed by Schick et al. (2023), Yao et al. (2022), and Mialon et al. (2023), can rely on specialized tools like calculators, code execution environments, and specific mathematical tools to carry out such operations seamlessly. Furthermore, according to the Chinchilla scaling law (Hoffmann et al., 2023), LLMs trained with more data will exhibit increased performance in a wide range of tasks, including those required to solve probability exercises like the one in question. Thus, we can expect an improved generation of large language models to significantly enhance their performance in solving such exercises in the near future.

The availability of intelligent assistants such as ChatGPT has revolutionized the way we approach teaching and learning. With the help of AI, students now have access to a vast collection of resources and receive personalized support that was previously impossible.

As a result, traditional teaching methods such as completing collections of exercises or web quizzes for solving numerical problems are becoming less relevant and less effective. However, LLMs present new possibilities: since students do not know a priori whether the answers provided by the chat are correct or incorrect, they must critically analyse and understand the answers, which improves their knowledge and enables them to work independently. With the help of ChatGPT, they can be asked to solve a series of statistical exercises and explain why an answer is correct or incorrect, which can force them to be more critical of the subject content.

Moreover, LLMs designed to interact in the form of dialog offer interesting opportunities to be used as tutors. Students can not only request to solve and explain an exercise, but also engage in a dialog with the LLM to request further explanations and interactively explore elements they do not fully understand. Moreover, LLMs can create new exercises of the same kind for students to practice and prepare better for exams, and they can correct the answers in a personalized way.

This shift towards personalized, AI-based learning has the potential to revolutionize education, making it more accessible and effective for students around the world. While traditional teaching methods may still have their place, it is clear that LLM powered intelligent assistants like ChatGPT are changing the way we learn and teach, paving the way for a more efficient and effective educational future.

The versatility of LLMs also opens the door to fine-tune models for specific use-cases in education, either through model fine-tuning or prompt engineering and in-context learning. However, it is important to be mindful of the existing limitations and risks associated with these models, including biases, inaccuracies, and the documented tendency to stray away harmfully from the intended context of operation.

# STATEMENTS & DECLARATIONS

## FUNDING

This work has been partially supported by grant TED2021-131295B-C33, funded by MCIN/AEI/10.13039/501100011033 and by the European Union NextGenerationEU/PRTR; and by grant XMIDAS, ref. PID2021-122640OB-I00, funded by the Spanish Ministry of Science and Innovation.

## CONFLICT OF INTEREST

The authors report no conflict of interest.

## CONTRIBUTIONS

AU: Conceptualization, Methodology, Formal analysis, Investigation, collecting data, analyzing data, Writing original draft ,AAA: Methodology ,exercise compilation and review, Writing – Review and editing,  IS: Methodology, Formal analysis, Writing – Review and editing  SH: exercise compilation and review, Writing – Review and editing,  MEC exercise compilation and review, Writing – Review and editing, RM: exercise compilation and review, Writing – Review and editing, ELC: exercise compilation and review, Writing – Review and editing